%% file: main.tex
\documentclass[10pt,twocolumn,letterpaper]{article}
\usepackage{cvpr}     
\usepackage{times}
\usepackage{color}
\usepackage{epsfig}
\usepackage{graphicx}
\usepackage{amsmath}
\usepackage{amssymb}
 \usepackage{mathrsfs}
\usepackage{booktabs}
\usepackage{comment}

\usepackage{bm} 
\usepackage{algorithm}
\usepackage{tabularx}
\usepackage{algpseudocode}
\usepackage{subfloat}
\usepackage{verbatim}
\usepackage[pagebackref,breaklinks,colorlinks]{hyperref}
\usepackage[capitalize]{cleveref}
\crefname{section}{Sec.}{Secs.}
\Crefname{section}{Section}{Sections}
\Crefname{table}{Table}{Tables}
\crefname{table}{Tab.}{Tabs.}

\usepackage{hyperref}
\hypersetup{
    colorlinks=true,            
    linkcolor=blue,             
    filecolor=magenta,          
    urlcolor=cyan,              
    pdftitle={Overleaf Example},
    pdfpagemode=FullScreen,
    }

\def\cvprPaperID{1921}
\def\confName{CVPR}
\def\confYear{2025}

\begin{document}


\title{Image Referenced Sketch Colorization Based on Animation Creation Workflow}

\vspace{-1em}
\author{
*Dingkun Yan\textsuperscript{1}
\qquad
*Xinrui Wang\textsuperscript{2}
\\
Zhuoru Li\textsuperscript{3}
\qquad
Suguru Saito\textsuperscript{1}
\qquad
Yusuke Iwasawa\textsuperscript{2}
\qquad
Yutaka Matsuo\textsuperscript{2}
\qquad
Jiaxian Guo\textsuperscript{2}
\\
\textsuperscript{1}Institute of Science Tokyo
\qquad
\textsuperscript{2}The University of Tokyo
\qquad
\textsuperscript{3}Project HAT}
\vspace{-1em}
\twocolumn[{%
    \renewcommand\twocolumn[1][]{#1}%
    \maketitle
    \begin{center}
        \centering
        \includegraphics[width=0.98\linewidth]{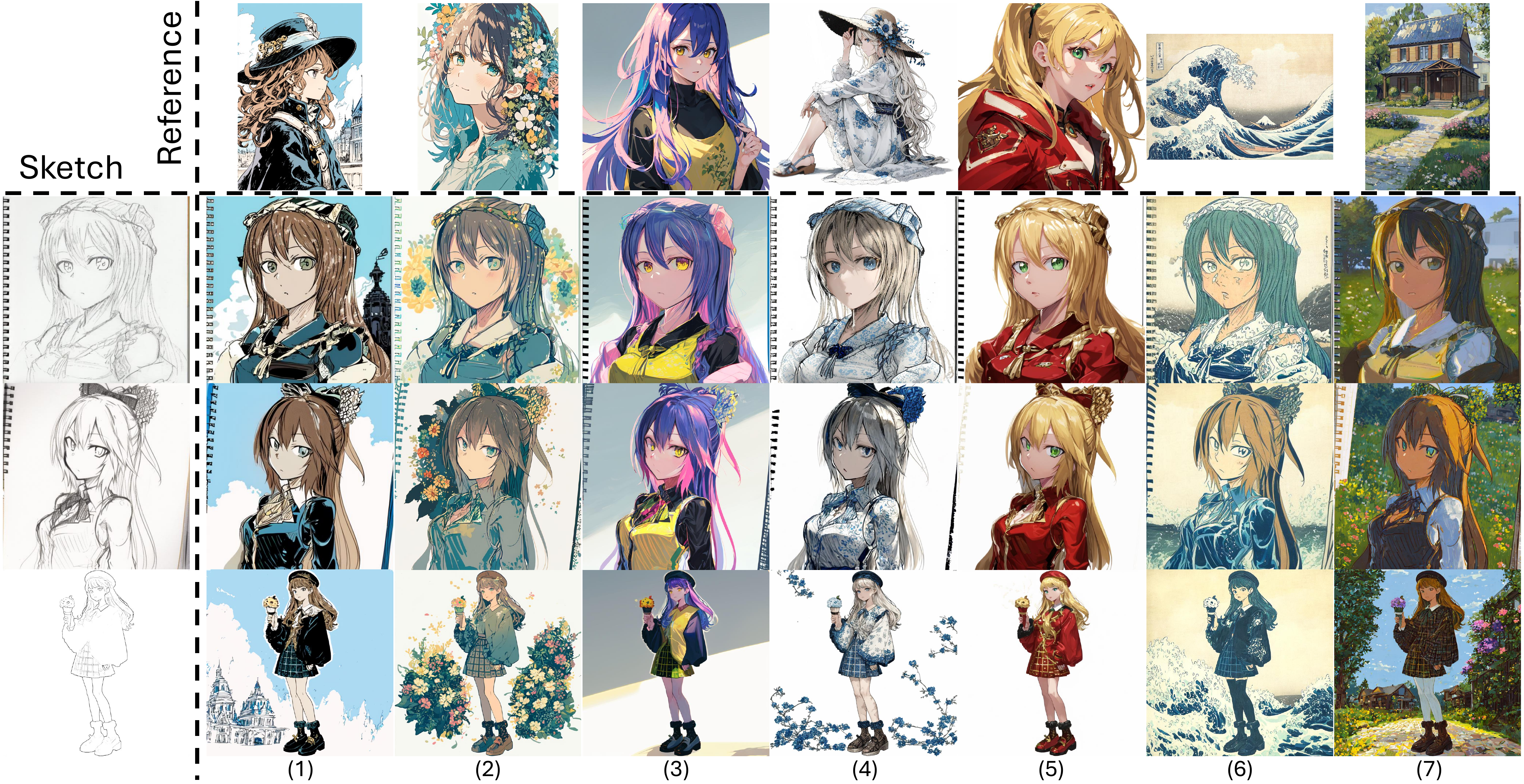}
        \vspace{-0.5em}
        \captionof{figure}{Given reference images, our proposed method automatically synthesizes high-quality sketch colorization results that loyally match the reference color distribution and are free from artifacts.} 
        \label{teaserfigure}
        \vspace{-0.5em}
    \end{center}%
    }]
\maketitle

\def\thefootnote{*}\footnotetext{Represent equal contribution to this work}\def\thefootnote{\arabic{footnote}}

\setcounter{figure}{2}
\begin{figure*}[t]
    \centering
    \includegraphics[width=\linewidth]{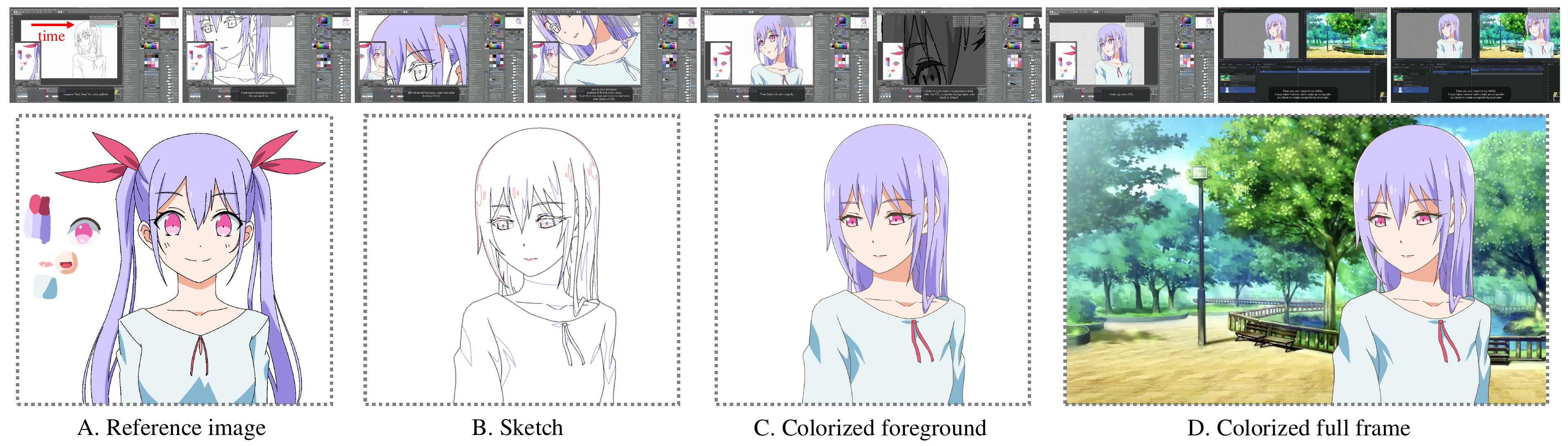}
    \vspace{-2em}
    \caption{Illustration of colorization workflow in professional animation studios. A: character designers design characters as references. B: Senior animators draw the sketches for the key frames. C: animators colorize the figures in the sketches according to the character designs, and D: animators colorize the background of the sketches and merge foreground and background into finished frames.}
    \label{pipeline}
    \vspace{-1em}
\end{figure*}
\setcounter{figure}{1}

\begin{abstract}
\vspace{-1em}
Sketch colorization plays an important role in animation and digital illustration production tasks. 
However, existing methods still meet problems in that text-guided methods fail to provide accurate color and style reference, hint-guided methods still involve manual operation, and image-referenced methods are prone to cause artifacts. To address these limitations, we propose a diffusion-based framework inspired by real-world animation production workflows. 
Our approach leverages the sketch as the spatial guidance and an RGB image as the color reference, and separately extracts foreground and background from the reference image with spatial masks. Particularly, we introduce a split cross-attention mechanism with LoRA (Low-Rank Adaptation) modules. They are trained separately with foreground and background regions to control the corresponding embeddings for keys and values in cross-attention. This design allows the diffusion model to integrate information from foreground and background independently, preventing interference and eliminating the spatial artifacts. During inference, we design switchable inference modes 
for diverse use scenarios by changing modules activated in the framework. Extensive qualitative and quantitative experiments, along with user studies, demonstrate our advantages over existing methods in generating high-qualigy artifact-free results with geometric mismatched references. Ablation studies further confirm the effectiveness of each component. Codes are available at \url{https://github.com/tellurion-kanata/colorizeDiffusion}.


\end{abstract}

\input{documents/introduction}

\input{documents/related}

\input{documents/method}
\input{documents/experiment}
\input{documents/conclusion}

\newpage

{\small
\bibliographystyle{ieee_fullname}
\bibliography{main}
}

\end{document}

%% file: documents/introduction.tex
\vspace{-1.5em}
\section{Introduction}
\vspace{-0.5em}
The past decades have witnessed the development of Animation as an artistic form, which has gained great popularity worldwide. The current workflow of animation creation is labor-intensive, and the growing demands of animation from the market are causing animation studios to fall short of hands, bringing severe problems to the industry.

\begin{figure}[t]
    \centering
    \includegraphics[width=\linewidth]{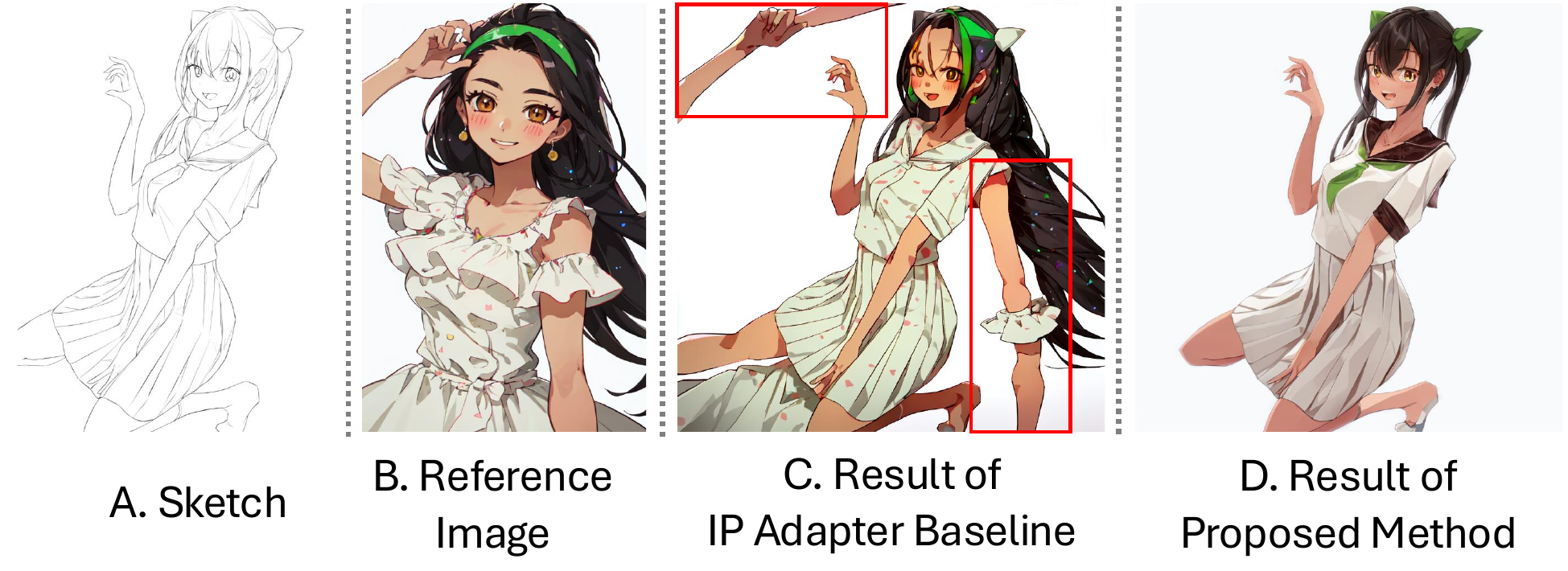}
    \vspace{-1.5em}
    \caption{Illustration of spatial entanglement. We use red rectangles to highlight the spatial entangled artifacts in the result of the IP-Adapter baseline, where additional arms appear unexpectedly, and the model mistakenly synthesizes long hair.}
    \label{entanglement}
    \vspace{-1.5em}
\end{figure}

The most manual labor-intensive procedure in animation production is sketch colorization, and animators working on colorization also take up the largest share among all employees in the animation production industry. To reduce the human labor needed and to automate the animation production, machine learning algorithms have been applied for sketch colorization \cite{ZhangLW0L18, LeeKLKCC20, zouSA2019sketchcolorization, li2022eliminating, zhang2021user}. However, current methods are still not optimal for real-world production pipelines: Text-based colorization methods\cite{controlnet-iccv, zabari2023diffusing} fail to provide accurate guidance on the color and style information of the images. User-guided methods \cite{zhang2021user, cho2023guiding} still involve manual operation in the process, making them less efficient. Image referenced methods \cite{yan2024colorizediffusion, animediffusion} can be seamlessly integrated into the current pipeline, but the spatial mismatches between reference images and sketches are causing severe artifacts and unexpected extra objects, which is termed as spatial entanglement in \cite{yan2024colorizediffusion} and shown in Figure \ref{entanglement}.

To build a sketch colorization framework that meets the requirements of the real-world animation production pipeline, we start with the observation of the manual sketch colorization workflow in real-world animation production. As is shown in Figure \ref{pipeline}, the workflow consists of the following key steps: Firstly, character designers design the characters as references. Secondly, senior animators draw the sketches for each frame. Thirdly, animators colorize the figures in the sketches according to the character designs. Finally, animators colorize the background of the sketches and finish the whole colorized frames. 

\setcounter{figure}{3}

Following our observation, we designed a diffusion-based framework to mimic the sketch colorization workflow step-by-step. To be smoothly integrated into the current production workflow, the proposed framework leverages a sketch image as the spatial reference and an RGB image as the color reference. Specifically, we use the high dimensional local tokens extracted by a feature extractor as reference embeddings to maintain the semantic information and adopt a multi-layer sketch encoder for precise spatial control of the background embeddings. 
To enable the separate colorization of foreground and background regions, we design a novel split cross-attention mechanism, where spatial masks are used to segment figures as foreground with the rest of the images as background, and corresponding LoRA weights are trained to modify the embeddings for keys and values within cross-attention layers. This design allows the diffusion model to integrate information from foreground and background independently, preventing interference and eliminating the need to adjust the well-trained backbone weights. During inference, we implement a switchable LoRA mechanism 
that provides precise control over the colorization process and enables different inference modes for various scenarios without changing model weights.


We train our model on 4.8M images and test on various scenarios. Experiments show that mimicking the real-world animation creation workflow yields several advantages. In qualitative analyses, our method synthesizes high-quality results loyally representing the color distribution of the reference images and free from artifacts and spatial entanglements. 
Quantitative comparisons also validate the superiority of the proposed method over existing methods by common criteria and benchmarks. What's more, user studies illustrate that artists prefer our method subjectively. 

In summary, our contributions are as follows: (1) We propose an image-referenced sketch colorization framework that can synthesize high-quality results free from artifacts and spatial entanglement by mimicking the animation production workflow. (2) We design a novel split cross-attention mechanism that enables the separate colorization of foreground and background in a single forward pass and the switchable LoRA module that allows users to switch colorization modes during inference. (3) Experiments show that our method outperforms existing methods in qualitative/quantitative comparisons and is preferred by human users in perceptive user study.

%% file: documents/related.tex
\section{Related Work}

\subsection{Latent Diffusion Models}

Diffusion Probabilistic Models \cite{HoJA20,0011SKKEP21} are a class of latent variable models inspired by nonequilibrium thermodynamics \cite{Sohl-DicksteinW15} and have achieved great success in image synthesis and editing. Compared to Generative Adversarial Networks (GANs)\cite{GoodfellowPMXWOCB14,KarrasLA19,KarrasLAHLA20,ChoiCKH0C18,ChoiUYH20}, Diffusion Models excel at generating highly realistic images with various contexts and able to be controlled by text prompts. However, the autoregressive denoising process of diffusion models, typically computed with a U-Net \cite{RonnebergerFB15} or a Diffusion Transformer (DiT) \cite{DiT,pixart}, incurs substantial computational costs. 

To address this limitation, Stable Diffusion (SD) \cite{RombachBLEO22,sdxl} as a class of Latent Diffusion Models (LDMs) was proposed, where a two-stage synthesis mechanism was adopted to enable diffusion/denoising process to be performed on a highly compressed latent space with a pair of pre-trained Variational Autoencoder (VAE), so as to significantly reduce computational costs. Concurrently, different researches of diffusion samplers have been conducted and proved to be effective in accelerating the denoising process \cite{SongME21,0011SKKEP21,0011ZB0L022,abs-2211-01095}. We adopt SD as our neural backbone, utilize the DPM++ solver \cite{abs-2211-01095,0011SKKEP21,KarrasAAL22} as the default sampler, and employ classifier-free guidance \cite{DhariwalN21,abs-2207-12598} to strengthen the reference-based performance.

\subsection{Image Prompted Diffusion Models}

 Currently, most diffusion models for image synthesis tasks are based on text prompts \cite{RombachBLEO22,sdxl, DiT,pixart}. However, there are tasks where text prompts can not provide enough information to precisely guide the image synthesis and editing, such as image-to-image translation \cite{KwonY23}, style transfer \cite{instantstyle, zhang2023inversion}, colorization \cite{animediffusion, yan2024colorizediffusion} and image composition \cite{zhang2023controlcom, kim2023reference}, and thus images are also used as prompts to provide reference information. The reference information extracted from prompt images varies from tasks: style transfer tasks adopt the textures and colors from reference images, image composition tasks focus more on the object-related information, and sketch colorization requires all the above.
 
There are two common practices to combine image prompts with diffusion models. Given image embedding vectors extracted by pre-trained feature extraction networks, existing methods either train an adapter module to inject the reference embedding vector into the backbone \cite{ip-adapter,t2i-adapter} or directly inject reference information into the backbone with attention layers \cite{hu2023animateanyone}. However, both of them \cite{ip-adapter,t2i-adapter,hu2023animateanyone} may introduce loss or mismatch of structure when inputs are not well paired, resulting in performance deterioration. In the sketch colorization task, specifically, these adapters provide conflicting spatial information from references and lead to unacceptable artifacts, as illustrated in Figure \ref{entanglement}.


\subsection{Sketch Colorization}

Sketch colorization has been a long-standing topic in computer vision. Interactive optimization-based method \cite{SykoraDC09} was employed for the task, and deep-learning-based methods \cite{ZhangLW0L18, KimJPY19, li2022eliminating, animediffusion, yan2024colorizediffusion} later became the mainstream due to the ability to synthesize high-quality and high-resolution images. There are three main technical solutions for deep learning methods: text-prompted sketch colorization \cite{KimJPY19,yan-cgf,controlnet-iccv}, user-guided sketch colorization \cite{ZhangLW0L18,s2pv5} and image reference sketch colorization \cite{li2022eliminating,yan-cgf,animediffusion}. User-guided methods can precisely colorize given sketches with detailed guidance from users, but the manual labor needed makes them unsuitable to be integrated into an automatic workflow. Text-prompted methods have received great popularity over recent years due to the development of Text-to-Image diffusion models, but it is challenging to precisely control colors, textures, and styles using text prompts. 
Image-referenced methods also benefit from the development of diffusion models, along with relevant works that enable image control \cite{controlnet-iccv,controllllite,controlnet-v11,t2i-adapter,ip-adapter}. However, the mismatch of reference images and sketches still results in severe deterioration. ColorizeDiffusion \cite{yan2024colorizediffusion} achieved notable progress in the colorization quality, yet it still suffers from spatial entanglement, which is shown in Figure \ref{entanglement}. In this paper, we base our method on the production pipeline of animation studios to use image references to guide colorization. We separate the foreground and background with an innovative switchable LoRA to improve colorization performance and prevent artifacts.

%% file: documents/method.tex
\vspace{-0.5em}
\section{Method}
\vspace{-0.5em}

\begin{figure*}
    \centering
    \includegraphics[width=1\linewidth]{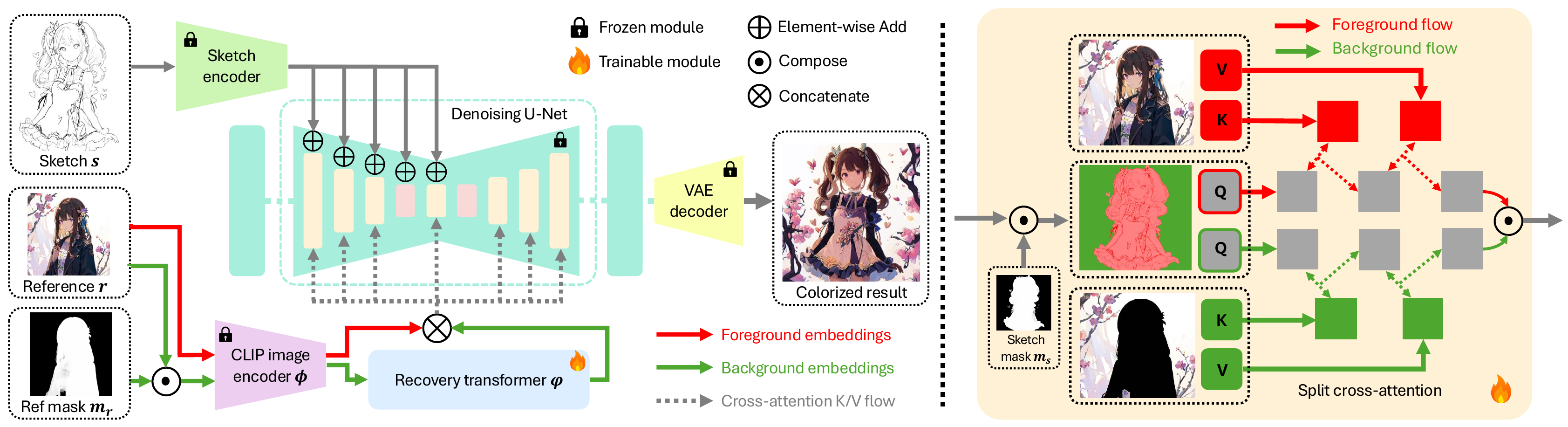}
    \vspace{-1.5em}
    \caption{Illustration of the proposed framework. We use reference masks to separate reference images into foreground and background and CLIP Image encoder $\phi$ to extract both regions into embeddings. The background embeddings first go through the recovery transformer $\varphi$ to recover detailed information, then concatenated with foreground embeddings as final K and V inputs for split cross-attention. Similar to Eq \ref{split-attention}, the compose operation is a spatial piece-wise function employed to separate foreground and background.}
    \label{framework}
\vspace{-1em}
\end{figure*}

Inspired by our observation of real-world animation production, we propose an image-referenced sketch colorization framework. As is shown in Figure \ref{framework}, it leverages a sketch image $\bm{X_s} \in \mathbb{R}^{w_s \times h_s \times 1}$, a reference image $\bm{X_r} \in \mathbb{R}^{w_r \times h_r \times c}$ and a foreground mask $\bm{X_m} \in \mathbb{R}^{w_s \times h_s \times 1}$ as inputs, and returns the colorized result $\bm{Y} \in \mathbb{R}^{w_s \times h_s \times c}$, with $w$, $h$ and $c$ representing the width, height and channel of the images. All components of the framework are based on the animation production workflow with the following design:

Firstly, to mimic the animator's colorizing sketches with character designs as references, we utilize a pre-trained vision transformer (ViT) to extract image embeddings as reference information. The embeddings are later injected into the diffusion backbone with split cross-attention layers. Secondly, to integrate the sketches into the framework, we adopt a multi-layer sketch encoder to inject sketch information into the latent layers of the diffusion backbone as spatial guidance. Thirdly, based on the behavior of animators separately colorizing the foreground and background of the sketch, we propose a novel split cross-attention mechanism that uses spatial masks to separate the trainable LoRA modules corresponding to the keys and values for foreground and background for training. A switchable LoRA mechanism is then applied during inference for different application scenarios with different colorization modes. The implementation details are described in the supplementary materials.

\subsection{Pretrain of the Diffusion Backbone}
The backbone of the proposed framework consists of a pre-trained VAE, a sketch encoder, a denoising U-Net, and a pre-trained Vision Transformer (ViT) functioning as the image encoder from OpenCLIP-H \cite{RadfordKHRGASAM21,openclip,openclip-2,schuhmann2022laionb}. 
We denote sketch images, reference images, and ground truth as $s$, $r$, and $y$, respectively. The VAE encoder, U-Net, and ViT are represented by $\mathcal{E}$, $\theta$, and $\phi$, respectively. The timestep $t$ starts from $T-1$ and goes to $0$, where $T$ is the diffusion steps, set to 1000. The training objective of the diffusion model is to denoise the intermediate noisy image $z_t$ via noise prediction:

\vspace{-1em}
\begin{equation}
    \mathcal{L}(\theta)=\mathbb{E}_{\mathcal{E}(y),\epsilon,t,s,r}[\|\epsilon-\epsilon_{\theta}(z_{t},t,s,\phi(r))\|^{2}_{2}].
\end{equation}
\vspace{-1em}


To pre-train the diffusion backbone, We initialize the VAE and U-Net with WaifuDiffusion \cite{waifudiffusion} and train networks with a dynamic reference drop rate decreasing from 80\% to 50\% as training progresses to avoid the distribution shift mentioned by \cite{yan2024colorizediffusion}. VAE, U-Net, and sketch encoder are then frozen during the training of the full framework. 

\subsection{Color Reference Extraction}

Following the real-world animation production workflow, we utilize images as color references for sketch colorization. The commonly used ViT-based image encoder networks have two kinds of output embeddings: the CLS embeddings $\bm{E_{cls}} \in \mathbb{R}^{bs \times 1 \times 1024}$ and the local embeddings $\bm{E_{local}} \in \mathbb{R}^{bs \times 256 \times 1024}$, where $bs$ represent the batch size. The CLS embeddings are projected to CLIP embedding space for image-text contrastive learning, with spatial information compressed and connected to text-level notions, and are employed as color or style references by previous image-guided methods \cite{ip-adapter, instantstyle}. 
ColorizeDiffusion \cite{yan2024colorizediffusion}, on the contrary, reveals that local embeddings also express text-level semantics, indicating that they express more details regarding textures, strokes, and styles, enabling the network to generate better reference-based results, especially for transferring detailed textures and strokes. Therefore, the proposed method follows \cite{yan2024colorizediffusion} to adopt local tokens as color reference inputs for the framework.

However, the excessive spatial information of image semantics and compositions contained in local embeddings leads to frequent occurrences of artifacts such as overflow of color regions and unexpected objects outside sketches. Illustrated in Figure \ref{entanglement}, such artifacts widely exist in frameworks with image references \cite{ip-adapter,instantstyle,t2i-adapter}. To eliminate this problem, 
we follow the real-world workflow to explicitly separate the foreground and background with spatial masks during colorization and describe the detail in \ref{split_cross_attention} and \ref{switchable_LoRA}.

\begin{figure}[t]
    \centering
    \includegraphics[width=1\linewidth]{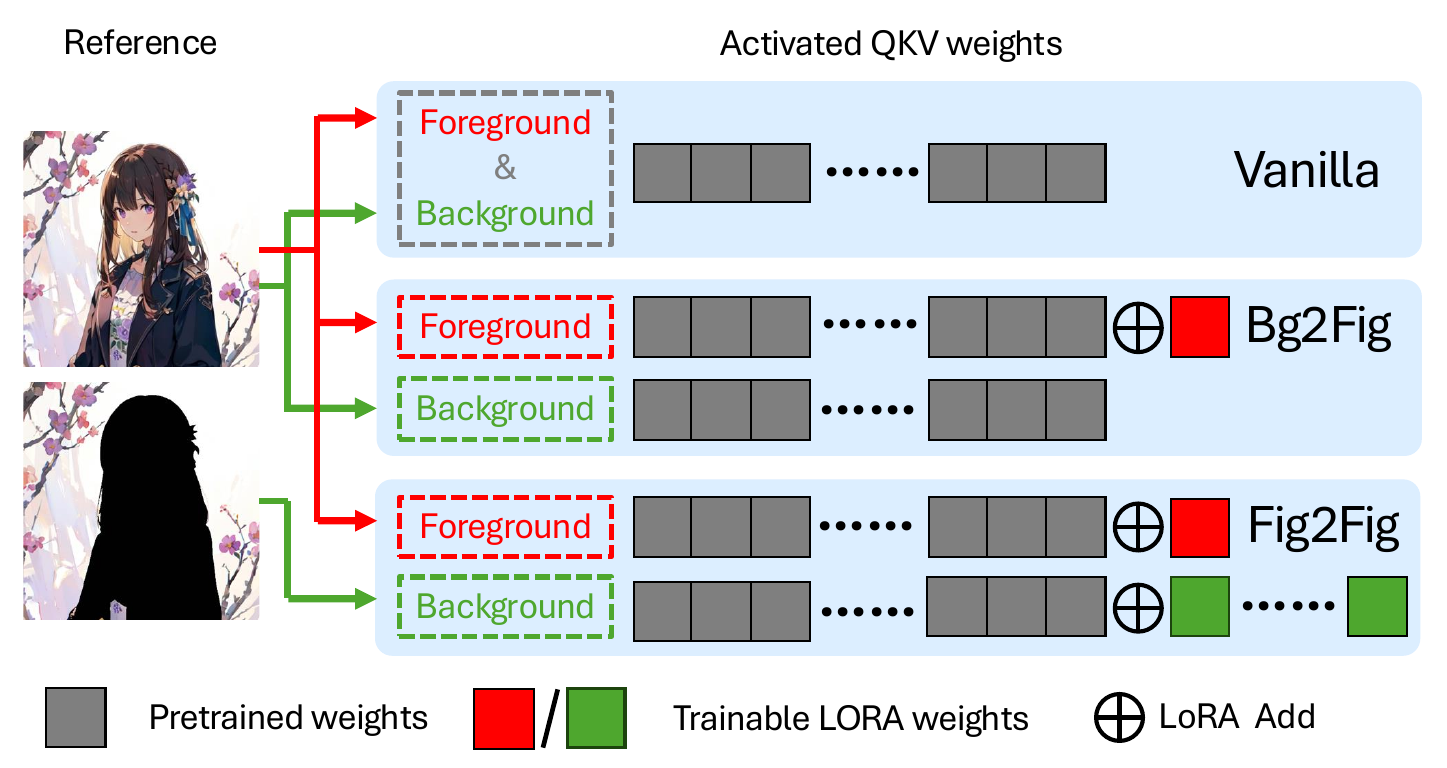}
    \vspace{-2em}
    \caption{Based on the LoRA weights, the proposed method can merge the foreground and background features in one forward pass and switch between three inference modes. We denote the dimension of pre-trained weights as CH. The rank of foreground LoRA is fixed at 16, while the rank of background LoRA is 0.5*CH.}
    \label{mode}
    \vspace{-1.5em}
\end{figure}

\subsection{Split Cross-Attention}
\label{split_cross_attention}

In anime images, the foreground regions and background regions differ distinctively in color distribution, color block sizes, tones, and textures. Thus, the colorization of foreground and background is separated into two independent steps in the animation production workflow. Following this scheme, we propose a novel split cross-attention mechanism to substitute the cross-attention layers in the diffusion backbone to separately process foreground and background regions with different parameters in a single forward pass.


A split cross-attention layer consists of two groups of trainable LoRA weights 
$W_{f}^{t}$ and $W_{b}^{t}$, which include query weights $W_{f}^{q}$ and $W_{b}^{q}$, key weights $W_{f}^{k}$ and $W_{b}^{k}$, and value weights $W_{f}^{v}$ and $W_{b}^{v}$ for foreground and background QKV projection respectively. An open-sourced animation image segmentation tool \cite{anime-segmentation} is used to automatically extract the foreground mask $m_{s}$ and $m_{r}$ of sketches and reference images. Regions with pixel values larger than thresholds $ts_{s}$ and $ts_{r}$ are considered as foreground, otherwise background. For foreground LoRAs, we set the ranks as 16; for background LoRA, the rank is formulated as $r=0.5*min(D_{q}, D_{kv})$, where $D_{q}$ and $D_{kv}$ are dimensions of queries and keys/values for the corresponding cross-attention layers.

As foreground and background regions of animation images feature different textures, color distribution, and color tones, injecting the embeddings of foreground and background directly into split cross-attention results in deterioration in structure preservation, synthesis quality, and stylization. 
Therefore, we further add a trainable recovery transformer $\varphi$ to process the background embeddings and facilitate better integration of the foreground and background reference information into the diffusion backbone.

We define query inputs (forward features) as $\bm{z}_{f}$, $\bm{z}_{b}$, key and value inputs (reference embeddings) as $\bm{e}$, $\bm{e}_{b}$, attention outputs as $\bm{y}$ in the following sections, where the index $f$ and $b$ indicate foreground and background respectively. Specifically, $\bm{e}$ denotes the reference embeddings extracted from the whole reference image $r$, formulated as $\bm{e}=\phi(\bm{r})$; and $\bm{e}_{b}=\varphi(\phi(\bm{r}_{b}))$, where $\bm{r}_{b}$ is the background region of the reference image. During training, the proposed split cross-attention can be formulated as follows:

\vspace{-1em}
\begin{equation}
\begin{small}
    \bm{y} = \begin{cases}
    \mathbf{Softmax}(\frac{(\hat{W}_{f}^{q}\bm{z}_{f})\cdot(\hat{W}_{f}^{k}\bm{e})}{d})(\hat{W}_{f}^{v}\bm{e}) & \text{if $\bm{m_{s}} > ts_{s}$}\\
    \mathbf{Softmax}(\frac{(\hat{W}_{b}^{q}\bm{z}_{b})\cdot(\hat{W}_{b}^{k}\bm{e}_{b})}{d})(\hat{W}_{b}^{v}\bm{e}_{b}) & \text{if $\bm{m_{s}}\leq ts_{s}$}
    \end{cases}
    \label{split-attention}
\end{small}
\end{equation}
\vspace{-0.5em}

where $\hat{W}_{f}^{t} = W^{t} + W_{f}^{t}$, and $W^{t}$ represents the pre-trained weights, which remain frozen during training. Similarly, $\hat{W}_{b}^{t}$ follows the same approach. 


\subsection{Switchable inference mode}
\label{switchable_LoRA}

The application scenarios and the sketch-reference combinations in the real world may be complicated. Also, naively separating the foreground and background regions for colorization degrades the quality of background synthesis, especially when reference images have severe semantic mismatches with the sketches or have complicated backgrounds. Therefore, we design three different inference modes: \textit{Vanilla}, \textit{Bg2Fig}, and \textit{Fig2Fig} for different scenarios based on the weights and reference inputs used for KV calculation. We visualize all inference modes in Figure \ref{mode}.

\textit{Vanilla} mode 
only utilizes the pre-trained weights for cross-attention modules, with $W_{f}^{t}$, $W_{b}^{t}$ and recovery transformers deactivated. It's suitable for most scenarios but suffers from spatial entanglement when references are figure-only images.

\textit{Bg2Fig} mode activates only the foreground-related LoRA weights $W_{f}^{t}$ during inference and is used when reference images are figure images with complicated backgrounds. This mode outperforms \textit{Vanilla} mode in character colorization and \textit{Fig2Fig} mode in background generation as its foreground weights are further optimized by LoRAs. 

\textit{Fig2Fig} is designed for figure-to-figure colorization, where reference images are figures with simple background composition. This mode activates both LoRA weights $W_{f}^{t}$, $W_{b}^{t}$, and the recovery transformer and uses spatial masks to separate foreground/background embeddings for calculating query and key/value inputs. 
It effectively eliminates the spatial entanglement in reference-based sketch colorization. 

%% file: documents/experiment.tex
\begin{figure}[t]
    \centering
    \includegraphics[width=1\linewidth]{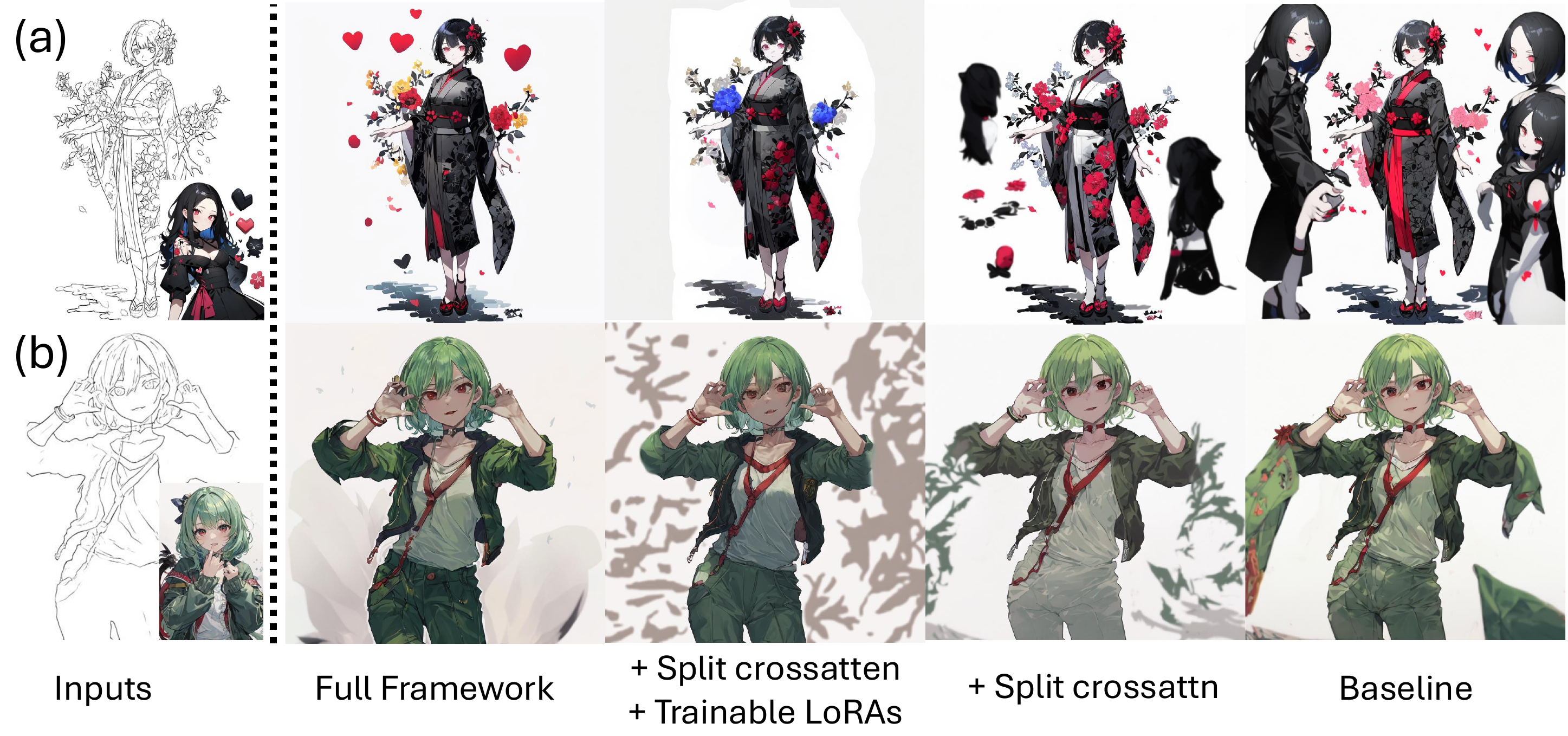}
    \vspace{-1.5em}
    \caption{
    Results of the ablation study. The baseline model demonstrates significant spatial entanglement; incorporating split cross-attention reduces artifacts, the trainable LoRAs improve color saturation and details, and the proposed complete pipeline produces high-quality results free of artifacts. Zoom in for details.}
    \label{ablation}
    \vspace{-1em}
\end{figure}

\section{Experiment}
\subsection{Implementation details}
We used Danbooru2021 dataset \cite{danbooru2021} to train and validate the proposed method. The training set contains over 4.8 million triples of (sketch, color, mask) images with various contents, and the validation set consists of 52,000 triples, with all the data excluded from the training set. Sketches were extracted by jointly using \cite{sketchKeras} and \cite{xiang2022adversarial}.
We first trained the denoising U-Net and sketch encoder using the dataset for 6 epochs and then froze the backbone and trained the recovery transformer and switchable LORAs on the dataset for 3 epochs. The training was conducted on 4x H100 (94GB) using Deepspeed ZeRO2 \cite{deepspeed} and the AdamW optimizer \cite{KingmaB14, LoshchilovH19} with the learning rate set to $0.0001$ and betas set to $(0.9, 0.999)$. Following \cite{yan2024colorizediffusion}, we dropped at least 50\% of the reference inputs in all training.

\subsection{Ablation study}
\noindent\textbf{Split cross-attention.} The proposed method aims to address spatial entanglement by simulating the animation workflow. To demonstrate the effectiveness of this workflow, we set up three frameworks: 1) baseline model without split cross attention, trainable LoRAs, and recovery transformer, 2) baseline model with split cross attention but no trainable LoRAs and recovery transformer, and 3) the proposed full framework. 

We show the qualitative comparison in Figure \ref{ablation} to validate the effectiveness of the proposed modules.
The baseline model causes severe spatial entanglement in generating additional figures in (a) and undesired clothes in (b). The application of split cross-attention mitigates the spatial entanglement but still causes artifacts and degrades the color saturation and details of the results. Collaborating split cross attention with trainable LoRAs improves the quality of results and further improves the background, but still suffers from artifacts. The proposed full framework enhanced by recovery transformers effectively eliminates spatial entanglement and synthesizes colorization results that have clear boundaries and rich details and textures, and loyally preserves the color distribution of reference images.

\begin{figure}[t]
    \centering
    \includegraphics[width=1\linewidth]{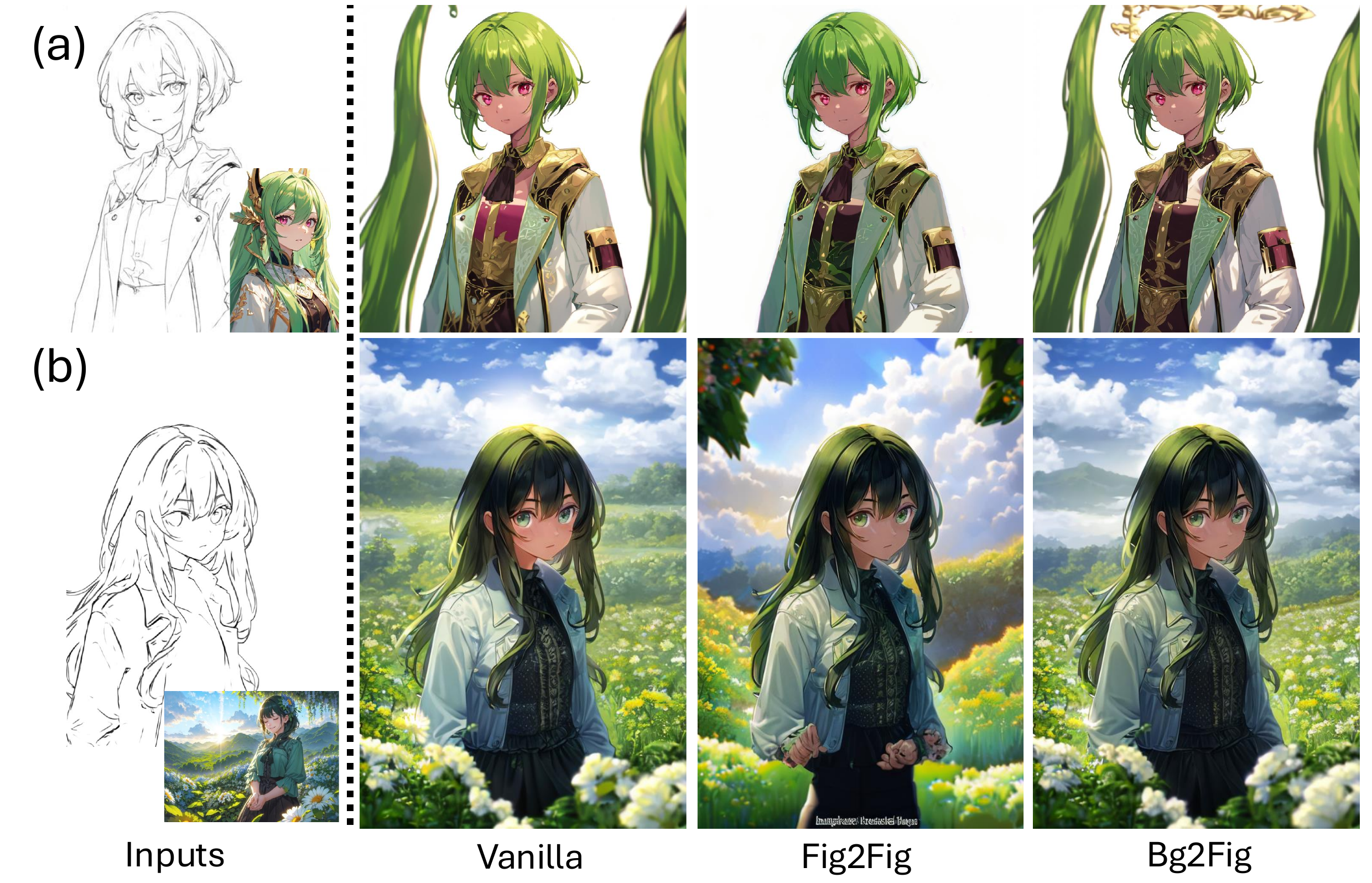}
    \vspace{-1.5em}
    \caption{Colorization results with three different inference modes. \textit{Fig2Fig} mode performs better in eliminating spatial entanglement, while \textit{Bg2Fig} and \textit{Vanilla} mode can generate vivid backgrounds and inpainting results.}
    \label{ablation2}
    \vspace{-1em}
\end{figure}

\noindent\textbf{Inference modes.} We illustrate the differences between inference modes and their use cases in Figure \ref{ablation2}. \textit{Fig2Fig} mode is fully mask-guided, enabling it to eliminate spatial entanglement shown in (a). However, it is less suitable for inpainting character sketches, as the region without the mask guide may suffer from the lack of reference information. \textit{Bg2Fig} performs similarly to the \textit{Vanilla} mode. With the help of foreground LoRA weight, the results demonstrate clearer segmentation and better stroke quality. All modes perform well for landscape sketches, which contain more structural detail than figure sketches, making them easier to colorize.

We show the comparison of local embeddings and CLS embeddings for \noindent\textbf{color reference extraction} in the supplementary materials.

\begin{figure*}[t]
    \centering
    \includegraphics[width=0.98\linewidth]{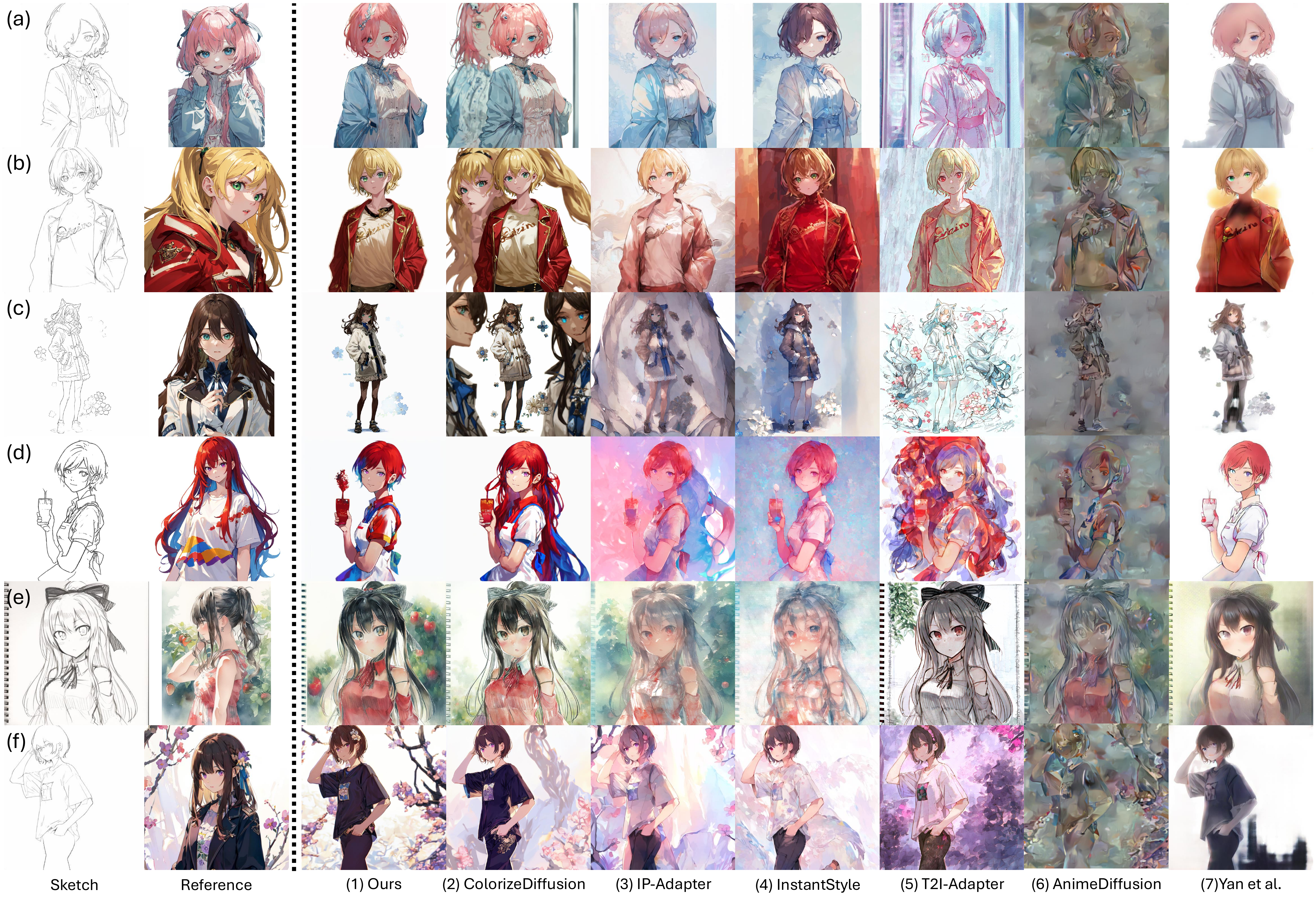}
     \vspace{-1em}
    \caption{Qualitative comparisons between our proposed method and existing methods show that our results are visually more appealing than those of image-prompt adapters \cite{ip-adapter, instantstyle, t2i-adapter} and the GAN-based method \cite{yan-cgf}. Compared to ColorizeDiffusion \cite{yan2024colorizediffusion}, the proposed framework eliminates spatial entanglement and improves overall quality.}
    \label{qualitative}
     \vspace{-1em}
\end{figure*}

\subsection{Comparison with baseline}

We compare our method with six existing reference-based sketch colorization methods \cite{yan-cgf,animediffusion,yan2024colorizediffusion} to demonstrate the superiority of the proposed framework.\\

\vspace{-0.5em}
\noindent\textbf{Baseline introduction.} Two baseline methods are the combination of SD \cite{RombachBLEO22}, ControlNet \cite{controlnet-iccv,controlnet-v11}, and IP-Adatper \cite{ip-adapter}. They adopt different cross-attention scales during denoising, labeled as \textit{IP-Adapter} and \textit{InstantStyle}, respectively. \textit{IP-adapter} baseline generates results with normal cross-attention scales, while \textit{InstantStyle} generates results using the ``style transfer'' weight type, which is claimed to prevent composition transfer by setting specific cross-attention scales to 0 according to \cite{instantstyle}. Following the official document of \cite{controlnet-v11}, we adopted Anything v3 \cite{anything}, a community-developed model, as the SD backbone for IP-Adapter-H and ControlNet\_lineart\_anime. All these models are officially implemented and claimed to be effective for anime-style image generation. The \textit{T2I-Adapter} baseline simply replaces IP-Adapter with T2I-Adapter-Style \cite{t2i-adapter,t2i-adapter-code}. Specifically, we introduced quality-related prompts and textual inversion, such as ``masterpiece'' and ``easynegative'' \cite{easynegative}, for T2I-model baselines to improve their image quality.\\

\vspace{-0.5em}
\noindent\textbf{Qualitative comparison.} We show the qualitative comparison of our proposed method and existing methods in Figure \ref{qualitative}, where most of the existing methods suffer from spatial entanglement in various cases. 
In rows (a)-(d), existing methods (2)-(6) failed to distinguish foreground and background regions and, therefore, synthesized artifacts in the background. Such artifacts become more obvious when reference images have complicated backgrounds in rows (e) and (f), where IP-Adapter and InstantStyle mixed reference composition with the sketch composition, making the generated results visually messy. The GAN-based method (7) successfully generated results with clear backgrounds, but the color preservation, texture transformation, and detail qualities are much poorer than diffusion-based methods. 

\begin{figure}[t]
    \centering
    \includegraphics[width=1\linewidth]{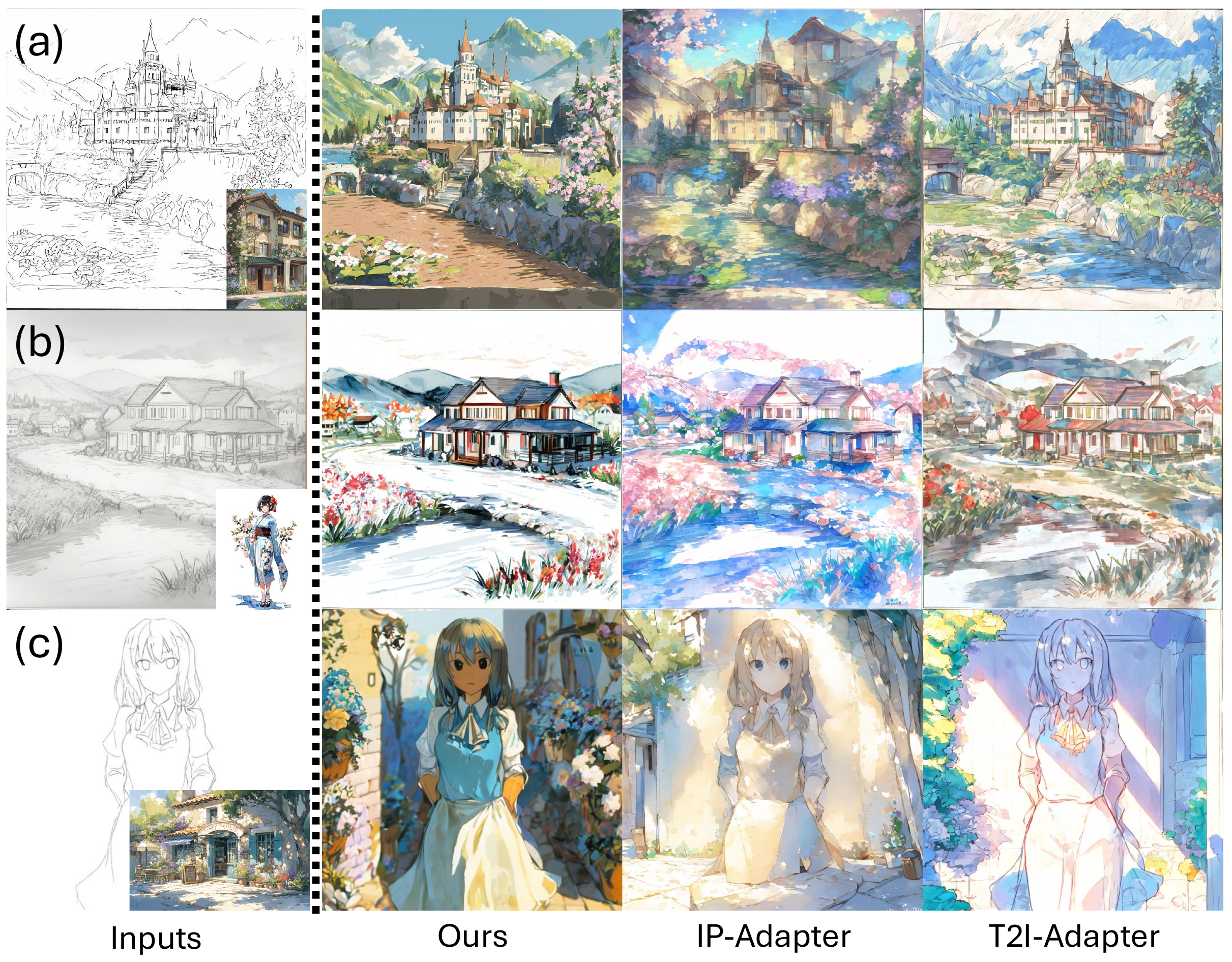}
    \vspace{-1em}
    \caption{Colorization results for semantically non-relevant sketch-reference pairs. Adapter-based methods fail to generate clear borders and high-quality textures, while our proposed method synthesizes visually satisfying results.}
    \label{qualitative2}
    \vspace{-1em}
\end{figure}

We also show the comparison of our proposed method and adapter-based methods on semantically non-relevant sketch-reference pairs in Figure \ref{qualitative2}. Both IP-Adapter and T2I-Adapter fail to separate foreground from background and colorize the whole image with the same tone and texture. 
Our proposed method, on the contrary, generated results with clear region boundaries, rich texture and details, and visually pleasant bright color, with the help of the proposed spatial aware split cross-attention mechanism and switchable LoRA. The qualitative comparisons demonstrate that our proposed method is effective for arbitrary input sketch-reference pairs and achieves high-quality colorization in various use scenarios.

\begin{table*}[t]
    \centering
    \caption{Quantitative comparison between the proposed model and baseline methods. We calculated 50K FID, 5K PSNR, 5K MS-SSIM, and 5K CLIP cosine similarity of image embeddings in this experiment. \dag: FID evaluation randomly selected color images as references, making it close to real-application scenarios. \ddag: Ground truth color images were deformed to obtain semantically paired and spatially similar references for evaluations.}
    \vspace{-0.5em}
    \begin{tabular}{|c|c|c|c|c|c|c|c|c|}
        \hline
        \multicolumn{5}{|c|}{Method} & {\dag 50K-FID $\downarrow$} & {\ddag PSNR$\uparrow$} & {\ddag MS-SSIM$\uparrow$} & {\ddag CLIP similarity$\uparrow$}\\
        \hline
        \multicolumn{5}{|c|}{Ours} & \textbf{6.8327} & 28.9144 & \textbf{0.6002} & \textbf{0.8829} \\
        \hline
        \multicolumn{5}{|c|}{\textit{ColorizeDiffusion} \cite{yan2024colorizediffusion}} & 9.5276 & 28.7384 & 0.5913 & 0.8775 \\
        \hline
	\multicolumn{5}{|c|}{\textit{IP-Adapter} \cite{controlnet-iccv,ip-adapter,anything}} & 38.9184 & 28.6767& 0.5478 & 0.8672 \\
        \hline
        \multicolumn{5}{|c|}{\textit{InstantStyle} \cite{controlnet-iccv,ip-adapter,instantstyle,anything}} & 40.8144 & 28.1090 & 0.4459 & 0.8042 \\
	\hline
        \multicolumn{5}{|c|}{\textit{T2I-Adapter} \cite{controlnet-iccv,t2i-adapter,anything}} & 41.1569 & 28.1275 & 0.3243 & 0.7180 \\
        \hline
        \multicolumn{5}{|c|}{\textit{AnimeDiffusion}\cite{animediffusion}} &  61.5999 & 27.8454 & 0.3185 & 0.7319\\
	\hline
        \multicolumn{5}{|c|}{Yan et al. \cite{yan-cgf}} & 27.0147 & \textbf{29.2483} & 0.5253 & 0.7634 \\
	\hline
	\end{tabular}
    \label{quantitative}
    \vspace{-1em}
\end{table*}

\noindent\textbf{Quantitative comparison.} Fréchet Inception Distance (FID) \cite{HeuselRUNH17} is a widely used quantitative creteria to evaluate the 
performance of image synthesis tasks. It calculates the perceptual distance of 
two distributions without requiring them to be semantically and spatially paired. 
We conduct a quantitative evaluation measured by FID on the entire validation set which includes 52k+ (sketch, reference) image pairs. Sketches are colorized with randomly selected reference images, ensuring each batch has semantically and spatially mismatched inputs.

Multi-scale structural similarity index measure (MS-SSIM), peak signal-to-noise ratio (PSNR), and CLIP score \cite{RadfordKHRGASAM21,openclip} assess the similarity between processed images and given ground truth. It requires the guiding reference to be aligned with the ground truth when applied to image-referenced sketch colorization. To fulfill this, we selected 5000 color images as ground truth, extracted the sketches from and used thin plate spline (TPS) transformation to spatially distort them as color references to build a test set of 5000 triples of sketch, reference, and ground truth, and evaluate all the 7 methods with MS-SSIM, PSNR and CLIP score on this dataset.  

We show the results of the quantitative comparison in Table \ref{quantitative}, where the proposed method significantly outperforms in FID, MM-SSIM, and CLIP similarity due to better texture and color synthesis. GAN-base method \cite{yan-cgf} achieved the best score in PSNR, with the proposed method ranked number 2. This is because the limited generation ability of \cite{yan-cgf} prevents it from generating complicated backgrounds and also hinders it from synthesizing bright colors and rich details of the figures. The close-to-average results make it advantageous to the calculation of PSNR \cite{blau2018perception}. 

\noindent\textbf{User study.}
The quality of sketch colorization is highly subjective and easily influenced by personal preference, we thus demonstrate how different individuals evaluate the proposed method and existing methods through an user study. 40 participants are invited to select the best results with two criteria: the overall colorization quality and the preservation of geometric structure of the sketches. We prepare 25 image sets and show each participant 16 image sets for evaluation.  
For each image set, participants are presented the colorization results of our proposed method as well as those generated by six existing methods. 

\begin{figure}[t]
    \centering
    \includegraphics[width=\linewidth]{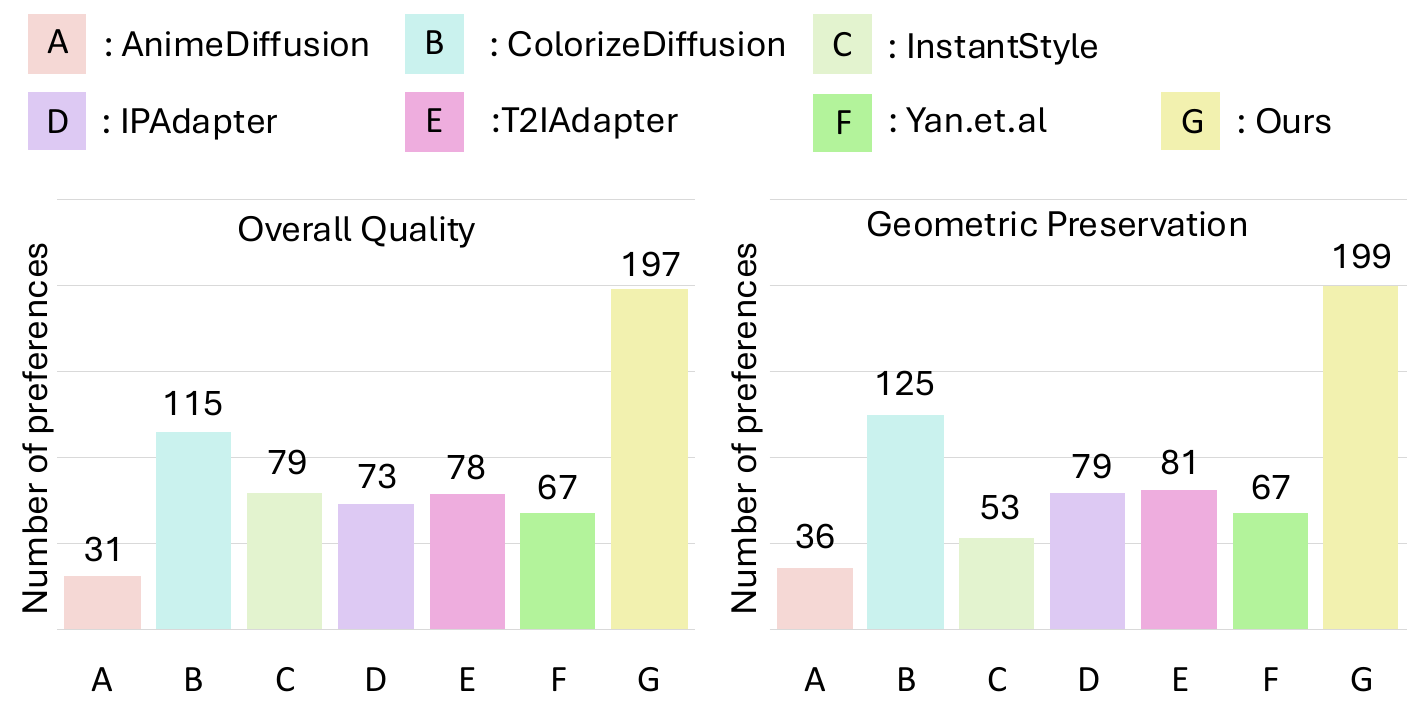}
    \vspace{-1em}
    \caption{Results of user study. Our method is preferred across all shown methods in overall quality and geometric preservation.}
    \label{userstudy}
    \vspace{-1em}
\end{figure}

The results of the user study are presented in Figure \ref{userstudy}, where our proposed method has received the most numbers of preferences among all the methods presented. To further validate the comparison, we applied the Kruskal-Wallis test as a statistical method. The results clearly demonstrate that the proposed method significantly outperforms all the existing methods in terms of user preference with a significance level of p \textless 0.01. All the images shown in the user study are included in the supplementary materials.

%% file: documents/conclusion.tex
\section{Conclusion}
In this paper, we proposed an image-guided sketch colorization framework inspired by real-world animation creation workflow. We implement a novel split cross-attention layer with spatial masks and corresponding LoRA weights that facilitates separate processing for foreground and background, eliminating the spatial entanglement, and also design a switchable LoRA mechanism that enables users to choose different inference modes for various use cases. 

However, the proposed methods strongly depend on the quality of extracted masks, and a corresponding failure case is given in the supplementary materials. Our future work focuses on further improving the similarity with references and extending the framework for video colorization.